\pdfoutput=1

\documentclass[11pt]{article}

\usepackage{array}
\usepackage{parskip}
\usepackage{booktabs}

\usepackage[]{acl}

\usepackage{times}
\usepackage{latexsym}

\usepackage[T1]{fontenc}

\usepackage[utf8]{inputenc}

\usepackage{microtype}
\usepackage{paralist}
\usepackage{graphicx}

\title{SalesBot: Transitioning from Chit-Chat to Task-Oriented Dialogues}

\author{Ssu Chiu$^\dag$\quad Maolin Li$^\star$\quad Yen-Ting Lin$^\dag$\quad Yun-Nung Chen$^\dag$\\
  $^\dag$National Taiwan University, Taipei, Taiwan \\
  $^\star$MediaTek Research, Cambridge, UK \\
  \small\texttt{r09944026@csie.ntu.edu.tw} \quad  \texttt{mmaolin.li@mtkresearch.com} \\
  \small \texttt{f08944064@csie.ntu.edu.tw} \quad \texttt{y.v.chen@ieee.org} \\ }

\begin{document}
\maketitle
\begin{abstract}
Dialogue systems are usually categorized into two types, open-domain and task-oriented. The first one focuses on chatting with users and making them engage in the conversations,
where selecting a proper topic to fit the dialogue context is essential for a successful dialogue. 
The other one focuses on a specific task instead of casual talks, e.g., finding a movie on Friday night, playing a song.
These two directions have been studied separately due to their different purposes.
However, how to smoothly transition from social chatting to task-oriented dialogues is important for triggering the business opportunities, and there is no any public data focusing on such scenarios.
Hence, this paper focuses on investigating the conversations starting from open-domain social chatting and then gradually transitioning to task-oriented purposes, and releases a large-scale dataset with detailed annotations for encouraging this research direction.
To achieve this goal, this paper proposes a framework to automatically generate many dialogues without human involvement, in which any powerful open-domain dialogue generation model can be easily leveraged.
The human evaluation shows that our generated dialogue data has a natural flow at a reasonable quality, showing that our released data has a great potential of guiding future research directions and commercial activities.
Furthermore, the released models allow researchers to automatically generate unlimited dialogues in the target scenarios, which can greatly benefit semi-supervised and unsupervised approaches.\footnote{{Our dataset, trained simulators, and annotations are available at: \url{https://github.com/MiuLab/SalesBot}.}}

\end{abstract}

\section{Introduction}
\begin{figure}[t!]
    \centering
    \includegraphics[width=\linewidth]{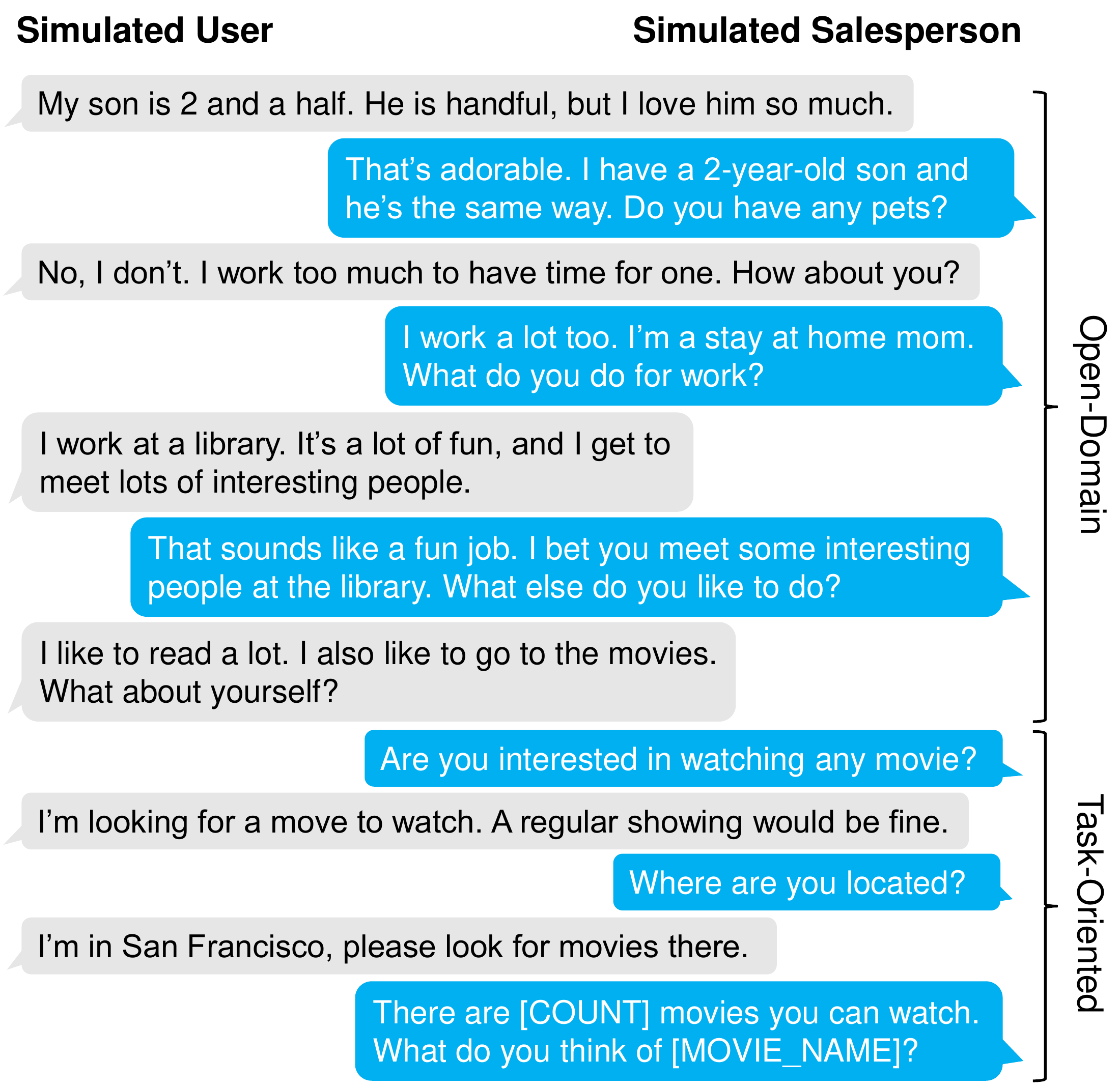}
    \vspace{-2mm}
    \caption{A generated example of smoothly transitioning from chit-chat to task-oriented dialogues. The sales discovers the user has a potential task-oriented intent related to watching movies and then decides to ask for the user's movie preference.
    }
    \label{fig:dial_example}
    \vspace{-3mm}
\end{figure}
Until now, researchers have often separated open-domain and task-oriented dialogues as two distinct types of tasks in the dialogue field. The publicly available datasets focuses on either open-domain or task-oriented dialogues.
For example, a lot of prior work focused on building open-domain dialogue systems~\cite{li-etal-2017-dailydialog,zhang2018personalizing,adiwardana2020humanlike}, which chat with users via suitable, engaging, safe conversations.
With the capability of pre-trained models, a large set of human conversations is adopted to train their capability of free chatting~\cite{zhang2020dialogpt,adiwardana2020towards,roller2021recipes}.
Although these models show the outstanding capability of communicating with human, they are not able to complete tasks as human assistants.
On the other hand, MultiWoz~\cite{budzianowski-etal-2018-multiwoz,NEURIPS2020_e9462095} and Schema-Guided Dialogue (SGD)~\cite{rastogi2020scalable} are two popular large-scale datasets of task-oriented dialogues, which include plenty of multi-domain dialogues with state information to track users' behaviors.
In task-oriented scenarios, the users have their goals before starting the conversations, so the way we evaluate the system's performance is whether the system can successfully complete the users' goals.

However, both skills of social chatting and task-oriented dialogues are important and may be used in a single conversation.

Considering that both skills are essential for a human-like dialogue system, the recent work \cite{sun-etal-2021-adding} merged those two capabilities by inserting chit-chat sentences into the existing task-oriented dialogue data.
The idea is to allow the agent gains more social, personalized communication skills when focusing on  task-oriented dialogue generation.
Even the released data contains both social and task-oriented dialogues, each dialogue still focuses on a task-oriented scenario where the user has the goal before starting the conversation.
 
In our target scenarios as illustrated in Figure~\ref{fig:dial_example}, the conversation starts without any specific goal in the user's mind, and the agent explores the potential task-oriented intents and smoothly transitions to a task-oriented conversation.
The focus of this paper is more similar to a salesperson's capability, where he/she needs to chat with the user and discovers the implicit task-oriented intents that fit the business purposes and navigates the user to complete a task, such as purchasing a product, reserving a restaurant, or booking a hotel room.
Hence, a new pipeline for constructing such data is proposed. 
Each dialogue in the released dataset starts with discovering a potential task-oriented intent of a user in the social conversation and ends in completing a specific task.
Even though high-quality chit-chats and task-oriented dialogues can be separately generated shown in prior work~\cite{NEURIPS2020_e9462095,adiwardana2020towards,roller2021recipes}, how to generate our desired dialogues has not been fully studied and remained unresolved.

\citet{yu2017learning} built a dialogue framework for users not having a clear intention, where mixing social responses into the conversation guides the flow to a specific movie they want to promote. 
Our paper has a similar idea about exploring the potential topics in the social conversations and then promoting the targeted tasks.
Although the prior work proposed the proper framework for the target scenarios, it required manual rules for dialogue strategies, making it difficult to scale.
Also, it only covers a single domain (movie) and there is no any publicly available data for following research work.
This paper covers more common topics by taking advantage of the existing natural language generation models trained on substantial dialogue datasets, and releases the first large-scale dialogue dataset with conversations naturally transitioning from chit-chats to task-oriented forms.
Our contributions can be summarized as 4-fold:
\begin{compactitem}
  \item We present a framework with a simulated user and a simulated salesperson to automatically generate dialogues that smoothly transitions from social chit-chats to task-oriented dialogues, where the components inside the framework can be easily replaced by any desired models for better flexibility. 
  \item Human evaluation on the generated dialogues demonstrates that the proposed method produces dialogues with reasonable quality and natural conversation flows.
  \item We release the \emph{first} large-scale dataset of dialogues transitioning from chit-chat to task-oriented scenarios, which contains the automatically generated dialogues and the detailed human annotations for enabling the future research work.
  \item The released framework with both user and sales simulators allows researchers to generate unlimited dialogues for semi-supervised and unsupervised usage.

\end{compactitem}

\section{Proposed Approach}
\begin{figure*}[t]
    \centering
    \includegraphics[width=\linewidth]{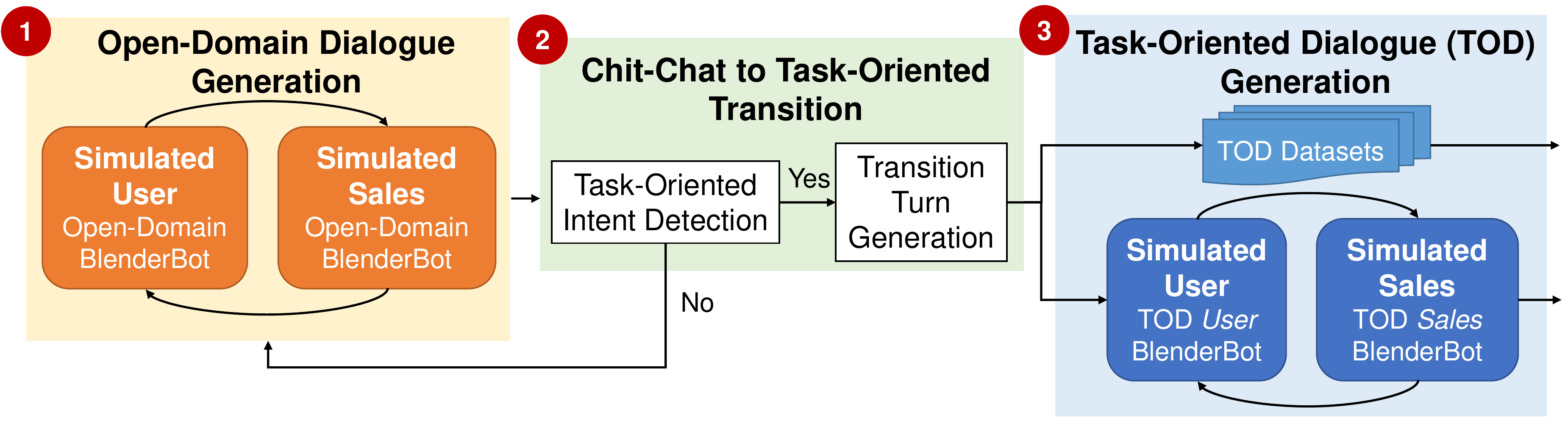}
    \vspace{-3mm}
    \caption{Illustration of the proposed framework that generates data transitioning from open-domain chit-chats to task-oriented dialogues.}
    \label{fig:pipeline}
    \vspace{-3mm}
\end{figure*}

Figure~\ref{fig:pipeline} illustrates our proposed framework for constructing the dataset. 
It can be divided into three main parts: (1) open-domain dialogue generation, (2) chit-chat to task-oriented transition, and (3) task-oriented dialogue (TOD) generation.

\subsection{Open-Domain Dialogue Generation}
\label{ssec:open-domain}
As shown in Figure~\ref{fig:dial_example}, the conversations start with social chatting between users and salespersons.
To generate high-quality open-domain dialogues, the pre-trained dialogue generation models can be adopted.
Here we choose BlenderBot \cite{roller2021recipes} as our pre-trained generation model due to its outstanding capability trained on the largest-ever open-domain data. It shows the ability to be engaging, knowledgeable, and empathetic at a certain level by multi-tasking on the Blended Skill Talk (BST) dataset \citep{smith-etal-2020-put} with several different datasets blending. 

Because users may explore any topics they want to discuss in a real-world setting, we manipulate the user and the sales to have different personas in order to cover wide-range topics in our generated dialogues.
This can be easily implemented by the package ParlAI\footnote{\url{https://parl.ai}} \citep{miller2017parlai}, which allows us to build two BlenderBots to self-chat with each other in order to construct various dialogues involving different personas~\cite{smith-etal-2020-put}.

\subsection{Chit-Chat to Task-Oriented Transition}
From a salesperson's perspective, \emph{how to capture the suitable timing} and \emph{how to promote the target products/tasks} are two main challenges.
This paper proposes two components to address the above issues; specifically, a task-oriented intent detector and a transition turn generator focus on capturing the suitable timing and deciding how to smoothly transition to the target task respectively.

\subsubsection{Task-Oriented Intent Detection}
\label{sssec:intent_detector}
To find out the good timing during social chatting, we focus on detecting whether the user currently has an implicit intent related to the target tasks.
In our case, an intent indicates what a user desires to do or what he/she is very likely to do if someone encourages him/her to do so. 
If our intent detector is able to capture any task-oriented intent in the social content with diverse topics, it tells us the suitable timing for guiding the dialogue to a specific topic and then transition to a corresponding task-oriented conversation.
Table~\ref{tab:intents} shows the intents we focus on in this paper, and other desired intents can be easily extended by our approach.

\begin{table}[t!]
\centering
\small
\begin{tabular}{p{2.35cm}p{4cm}}
\toprule
\textbf{Intent} & \textbf{Description} \\ 
\midrule
\textsf{FindMovies}          &   find movies to watch               \\
\textsf{GetTimesForMovie}    &   obtain the available time for watching a movie               \\
\textsf{FindAttractions}     &   find attractions to visit                      \\
\textsf{LookupMusic}         &   find music to listen to\\
\textsf{PlaySong}            &   play songs \\
\textsf{LookupSong}          &   find songs to listen to   \\
\bottomrule
\end{tabular}
\vspace{-2mm}
\caption{Descriptions of intents.}
\label{tab:intents}
\vspace{-3mm}
\end{table}

Although detecting intents in task-oriented dialogues has been studied for long time, the intent detection models trained on task-oriented datasets cannot be directly utilized.
The reason is that the intents in our scenarios are different from the intents in classical task-oriented data, where former ones are more \emph{implicit} and the latter ones are more \emph{explicit}.
For example, a user utterance with the intent \textsf{FindAttraction} in our case may be ``{\it I never visit France, but I heard that it is a good place.}'' instead of ``{\it Find me the landmarks in Paris.}'' in classical task-oriented dialogue datasets.
Therefore, this paper proposes to leverage the powerful capability of question answering (QA) systems to identify the potential task-oriented intents in a zero-shot fashion~\cite{Namazifar:2020}.
Specifically, we use the pre-trained QA model and ask whether the user has a certain intent given the current dialogue.
The questions need to be designed for describing the target task-oriented intents, and we use the following ways to create the questions focusing on task-oriented intents.\footnote{The manually-designed questions are listed in the Appendix \ref{sec:questions}.}
\begin{compactenum}
\item \textbf{Questions based on descriptions:} we create questions associated with all intents based on their natural language descriptions, e.g. ``\textit{Is the intent asking about playing songs?}'' for the intent \textsf{PlaySong}.
\item \textbf{Paraphrased questions:} to enhance the detection recall for open-domain dialogues, for each intent, we paraphrase the description-based questions via a high-quality paraphrasing T5 model  pre-trained on Quora Question Pairs data for its paraphrasing capability \cite{wang2017bilateral}.
\end{compactenum}

\begin{figure}[t]
    \centering
    \includegraphics[width=\linewidth]{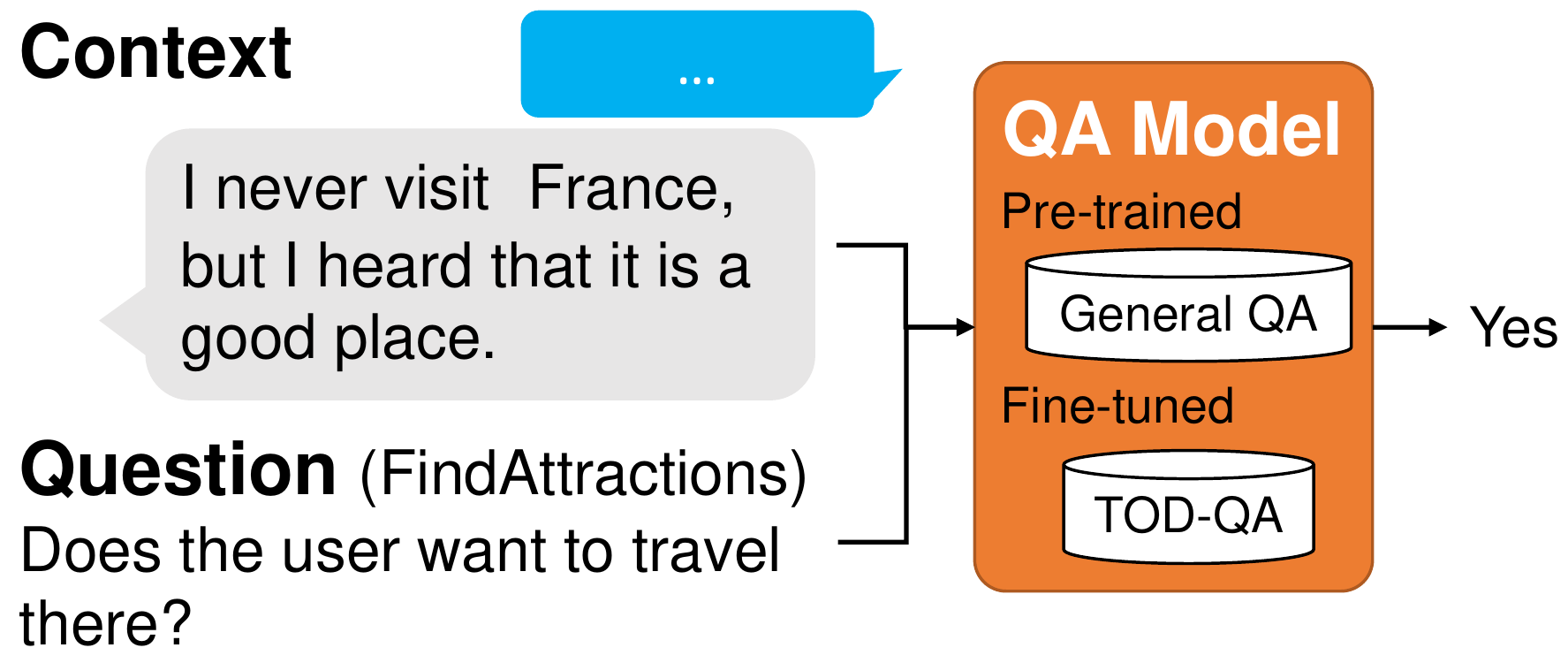}
    \vspace{-2mm}
    \caption{Zero-shot task-oriented intent detection.}
    \label{fig:intent}
    \vspace{-3mm}
\end{figure}

The proposed intent detector is illustrated in Figure~\ref{fig:intent}, where the inputs are the open-domain conversation along with intent-related questions, and the outputs are Yes/No answers to these questions.
We assume that a user has a task-oriented intent when the detector outputs Yes to the associated question.
Note that any type of QA models can be adopted in our framework.
Here we start with a QA model pre-trained on large open-domain QA data, e.g., SQuAD \citep{rajpurkar-etal-2018-know} or CommonsenseQA~\cite{talmor2019commonsenseqa}, which is supposed to be equipped with certain common knowledge and the reasoning ability useful for our intent detector. 
Furthermore, the general QA model may not be capable of correctly answering intent-related questions since the contexts and questions differ a lot from ones in the general QA data.
To reduce the mismatch, we fine-tune the QA model on a publicly available task-oriented dataset (e.g., SGD).
Specifically, the annotated intents in task-oriented dialogues are utilized to create the associated QA data, where there is a ground truth answer (Yes/No) to each intent-related question at all dialogue turns. Then the built training data (TOD-QA shown in Figure~\ref{fig:intent}) allows the general QA model to better identify task-oriented intents.
Although fine-tuned on the task-oriented dataset, we find that the model benefits from pre-training and thus it can be well applied to open-domain dialogues. 

\begin{figure}[t]
\small
\centering
\begin{tabular} { | p{1cm} p{5.8cm} | }
\hline
\multicolumn{2}{|c|}{\bf Template-based generation}\\
 \textbf{User:} & I like to read a lot. I also like to go to the movies. What about yourself? - \textbf{\underline{FindMovies}} \\
 \textbf{Sales:} & \emph{Do you want to find movies by genre and optionally director?}  \\ 
 \textbf{User:} & I'm looking for a movie to watch. A regular showing would be fine. \\
\hline
\end{tabular}
\vspace{2.5mm}
\begin{tabular} { | p{1cm} p{5.8cm} | }
 \hline
 \multicolumn{2}{|c|}{\bf Generative-based Re-generation}\\
 \textbf{User:} & I like to read a lot. I also like to go to the movies. What about yourself?\\ 
 \textbf{Sales:} & \underline{\it Are you interested in watching any movie?}  \\
 \textbf{User:} & I'm looking for a movie to watch. A regular showing would be fine. \\ 
 \hline
\end{tabular}
\vspace{-3mm}
\caption{The upper block is a template-based transition example. When detecting the task-oriented intent \textsf{FindMovies}, its intent descriptions trigger a template transition sentence (in \textit{italics}), and then these two user turns are used to re-generate a transition as shown in the lower block to substitute the template transition.}
\label{fig:transition_example}
\vspace{-3mm}
\end{figure}

\subsubsection{Transition Turn Generation}
\label{sssec:transition}
This section describes how we generate the transition turn that bridges open-domain and task-oriented dialogues.
Our transition turn generation procedure is composed of two parts: 1) using a template transition sentence to trigger the corresponding task-oriented user reaction and 2) re-generating the transition turn for better fluency and diversity.

\vspace{-1.5mm}
\paragraph{Template-based}
For each task-oriented intent, we adapt its intent description in the ontology to create a corresponding template question (e.g., \textit{Do you want to [Intent Description]?}) as the transition sentence shown in the upper block of Figure \ref{fig:transition_example}.
Although using template-based transition is simple and effective, it however makes the salesperson too aggressive and invariant to be professional. 

\vspace{-1.5mm}
\paragraph{Generative-based} 
To improve the fluency of transition and increase the diversity of word usage, we propose a generative-based approach to re-generate more smooth and nature transitions.
With a similar idea as \citep{ennen2021universal, sevegnani-etal-2021-otters}, our goal is to predict a transition utterance that can naturally bridge the past and the future utterances as below.
$$
p(a_t\mid u_t, u_{t+1}) = \prod_{k=0} p(a_{t,k}\mid u_t, u_{t+1}, a_{t,1:k-1}),\nonumber
$$
where $a_t$ is the current utterance, $u_t$ is the past utterance, $u_{t+1}$ is the future utterance, and $k$ the $k$-th token in $a_t$.

Specifically, we feed the last user's open-domain utterance and the first user's task-oriented utterance in our generated data as inputs, and learn to predict the template transition turn.
To learn the capability of connecting different topics smoothly, the newly published data OTTers~\cite{sevegnani-etal-2021-otters} is leveraged for training our generative model.
This data focuses on bridging two different topics via the transition in an entity path of a commonsense knowledge graph.
The assumption of using this dataset is that open-domain utterances can be viewed as the previous topic and task-oriented utterances as the new one, so learning the transition turn is the same as learning how to smoothly transition from open-domain to task-oriented dialogues.

\subsection{Task-Oriented Dialogue Generation}
\label{ssec:dialogue_generation}
After detecting the potential task-oriented intent and generating the transition turn, it is natural to continue the dialogue in a task-oriented scenario illustrated in the right part of Figure~\ref{fig:pipeline}.
Here we propose two ways of generating task-oriented dialogues following the transition turn.

\vspace{-1.5mm}
\paragraph{Merge SGD}
It is naive to simply merge an appropriate task-oriented dialogue taken from TOD data with a chit-chat dialogue to create such dialogue. 
In more details, all task-oriented dialogues in the SGD dataset are grouped by intents, and one TOD dialogue is sampled based on the detected task-oriented intent to append to the transition turn and form a new dialogue containing both chit-chat and TOD.
Note that the delexicalized version of SGD~\cite{sun-etal-2021-adding} is used to avoid severe inconsistency between open-domain and task-oriented parts.

\vspace{-1.5mm}
\paragraph{Task-Oriented Simulation} 
Different from open-domain social chatting, the roles in task-oriented dialogues are important.
Therefore, two task-oriented simulators are trained, one for users and another for salespersons.
Considering that training on task-oriented dialogues from scratch may limit the diversity of the generated dialogues, to generate the context-aware, fluent, and consistent conversations, we use the same type of open-domain dialogue generation models, BlenderBot~\cite{roller2021recipes}, and additionally train on either user turns or agent turns in task-oriented dialogues for TOD User BlenderBot and TOD Sales BlenderBot.
By allowing two simulators to talk with each other, 
they can generate endless conversations until one of the termination conditions is satisfied.
There are three commonly used termination strategies we use when building our dataset: 
(1) Any pre-defined keyword appears in the utterance, e.g., \textit{bye}. (2) The sales simulator generates a special token representing the ending of a dialogue. (3) When the dialogue starts to repeat itself, i.e., repeatedly producing the same utterances, because it usually means no more useful information.

The proposed framework enables us to construct a large-scale dataset with dialogues transitioning from open-domain to task-oriented scenarios, which align well with the salesperson's business potential.

\section{Data Quality Evaluation}
We use a widely-used crowdsourcing platform, Amazon Mechanical Turk (AMT)\footnote{\url{https://www.mturk.com/}}, to collect human feedback for our generated dialogues.

\subsection{Model Settings}
\label{ssec:model_settings}
\paragraph{Intent Detector} Our QA model is DistillBert \citep{sanh2020distilbert} pre-trained on the general QA data, SQuAD 2.0 \citep{rajpurkar-etal-2018-know}, and then fine-tuned on TOD data, SGD. 
The value of learning rate and batch size are 3e-5 and 64 respectively with AdamW optimizer \cite{loshchilov2018decoupled} for 20 epochs.

\vspace{-1.5mm}
\paragraph{Transition} The T5 (T5-small) model is trained to generate transitions with a learning rate of 5e-5 with Adafactor optimizer \cite{DBLP:journals/corr/abs-1804-04235} and batch size of 16. We train our models for 5 epochs and select the model with lowest loss in the dev set. 
During decoding, we mix top-K sampling of 80 and top-p (nucleus) sampling of 0.95~\cite{holtzman2020curious}.

\vspace{-1.5mm}
\paragraph{Dialogue Generation} To generate task-oriented utterances, we train our two simulators on the model BlenderBot-400M\footnote{\url{https://parl.ai/projects/blenderbot2/}} with the same training setting with generative transition.
During generation, we use top-K sampling \cite{fan-etal-2018-hierarchical}, where K is set to 120 for encouraging more diversity in dialogues. 

\begin{table}[t]
\small
    \begin{tabular} { | p{0.7cm} p{6.0cm} | }
     \hline
     \textbf{Sales:} & Cigarettes are pretty bad for you, but they're so tasty! Do you like to smoke in your free time?\\ 
     \textbf{User:} & Yes, I enjoy smoking in my free time. I also enjoy lifting weights and squatting. \\ 
     \textbf{Sales:} & That's great! I like to lift weights as well. It's a great way to stay in shape.\\ 
     \textbf{User:} & I love to play video games. I play a lot of League of Legends. What about you? \\ 
     \textbf{Sales:} & Yes, it is. I enjoy it a lot. What else do you enjoy doing in your spare time? \\ 
     \textbf{User:} & I enjoy playing video games as well! I also like to go to the movies.\\ 
     \textbf{Sales:} & \textit{I usually watch a movie. I found [count] movies. What about [title], [title], or [title]? - [Transition]}\\ 
     \textbf{User:} & I wanna find some Romance movies online.\\ 
     \textbf{Sales:} & What do you think about [title], [title] or [title]? \\ 
     \textbf{User:} & Is there anything else? Violent movies would be nice. I like the director Jia Zhangke. \\ 
     \textbf{Sales:} & There's [count] movie that might suit your interests. What about [title]? \\ 
     \textbf{User:} & Ash Is Purest White sounds great. I wanna watch that movie now. I wanna watch it with subtitles. \\ 
     \hline
    \end{tabular}
    \vspace{-1mm}
\caption{A partial dialogue sample presented to AMT workers, where the transition turn in \textit{italics} only highlighted  in Task 2.  \label{tab:AMT_example}}
\vspace{-2mm}
\end{table}

\subsection{Crowdsourcing Tasks}
\label{ssec:crowd_tasks}
We randomly pick about 4,000 dialogues for two human evaluation tasks submit to AMT. The first task is designed for collecting feedback about the \emph{entire} dialogue, while the second one focuses on the \emph{transition} part due to the main goal of this paper. 
Table~\ref{tab:AMT_example} shows the dialogue example presented to crowdworkers for evaluation.
Because our target dialogues can be viewed as salespersons' capability, we describe the context to the recruited crowdworkers that the given dialogues are between a beginner salesperson and his/her customer and ask workers to provide feedback from different aspects in terms of the sales' strategies.
Note that the annotators are not aware that the dialogues are machine-generated, so the collected feedback is for human sales communication skills.
Each task is briefly described below, and the full annotation guideline can be found in the Appendix \ref{sec:guideline}.

\vspace{-1.5mm}
\paragraph{Task 1: Salesperson-Customer Conversation} The workers were presented with one entire dialogue and asked to rate (from 1 to 5) the entire conversation from three aspects: \textbf{Relevance} (Q1---How relevant is the recommended product or service to the conversation context?), \textbf{Aggressiveness} (Q2---How aggressive is the salesperson's communication strategy?), and \textbf{Overall} (Q3---Do you think the sales conversation is overall a good example of making a sales recommendations?).

\vspace{-1.5mm}
\paragraph{Task 2: Chit-Chat to Task-Oriented Transition} In addition to the entire dialogue, we also explicitly highlight the transition turn in the dialogue when presenting to crowdworkers. Similarly to the first task but only focusing on the transition part, we asked workers to rate from 1 to 5 from the following aspects: \textbf{Right Time} (Q1---Is it a good timing to make the transition?), \textbf{Relevance} (Q2---Is the transition relevant to the conversation context?), \textbf{Aggressiveness} (Q3---Is the transition aggressive?), and \textbf{Overall} (Q4---Do you think it is overall a good transition?). 
In each question, the detailed descriptions of all ratings are given to crowdworkers to ensure they have consistent understanding for all ratings. 
In addition, to enrich the transition turns and ensure their quality, we generate 4 additional transitions and ask workers to choose the best one.
All transitions and ratings are included in our released data.

\paragraph{Task 3: Customer's Implicit Intent} 
Considering that detecting potential intents plays an important role in our framework, we further investigate the influence of intent detectors.
To evaluate the performance of different detectors, crowdworkers are presented with a conversation snippet and the detected intent results from three detectors, and they are asked to rank the intents in terms of their relevance to the conversation.
Three evaluated detectors are: \textbf{Detector1}---pre-trained on SQuAD 2.0 (Section~\ref{ssec:model_settings}), \textbf{Detector2}---additionally pre-trained on SWAG~\cite{zellers-etal-2018-swag} and CommonsenseQA~\cite{talmor2019commonsenseqa}, and \textbf{Detector3}---adapted from TransferQA \cite{lin-etal-2021-zero}, which learns dialogue state tracking knowledge from several general QA datasets. 
We evaluate 1,500 conversation snippets, and three workers are recruited to rank intents for each snippet.

\begin{figure*}[t]
\centering
    \begin{minipage}[b]{0.49\linewidth}
    \includegraphics[trim=1cm 1cm 0.5cm 1cm, clip=true,width=1.1\linewidth]{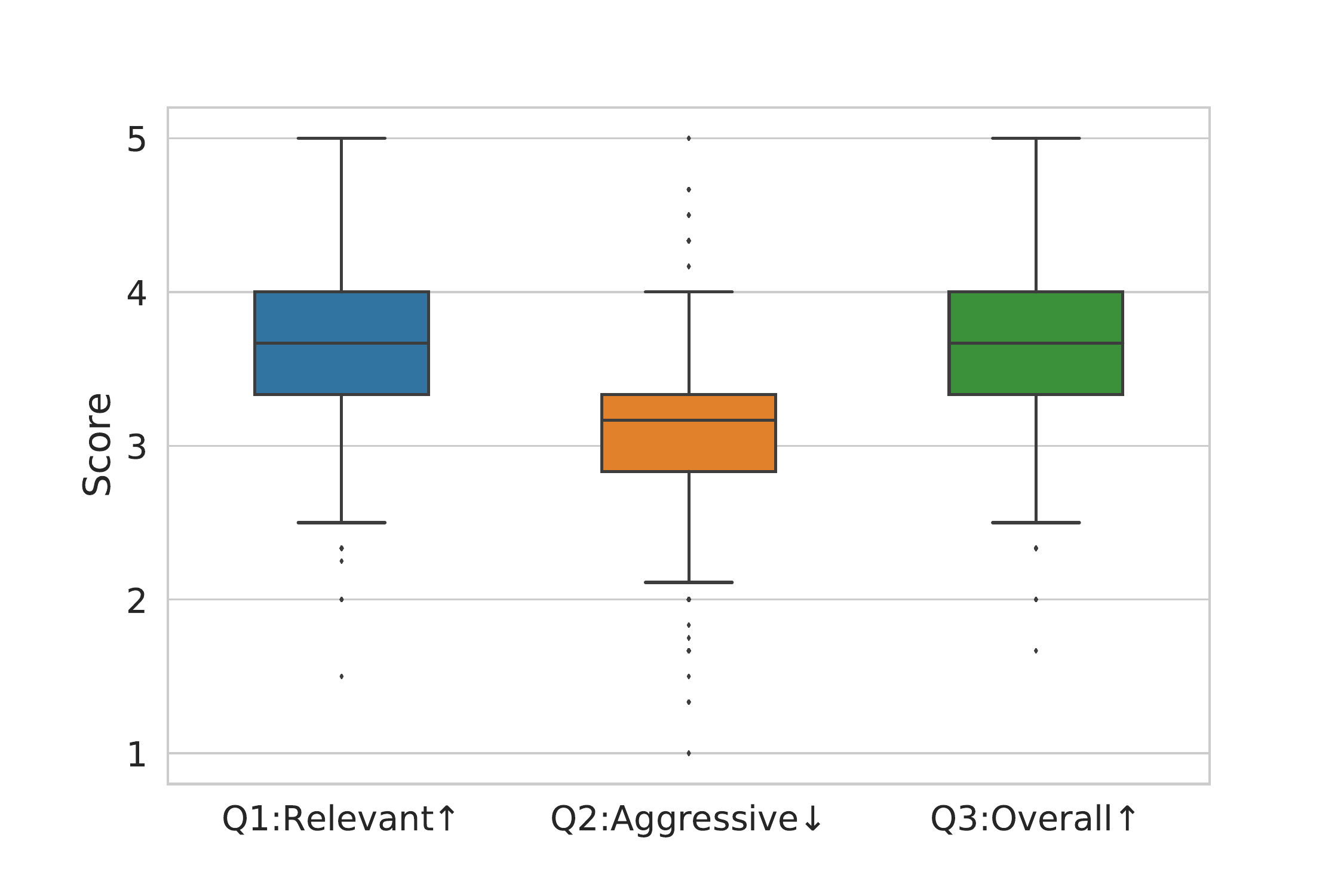}
    \vspace{-2mm}
    \end{minipage}
    \begin{minipage}[b]{.49\linewidth}
    \includegraphics[trim=1cm 1cm 0.5cm 1.5cm,clip=true,width=1.1\linewidth]{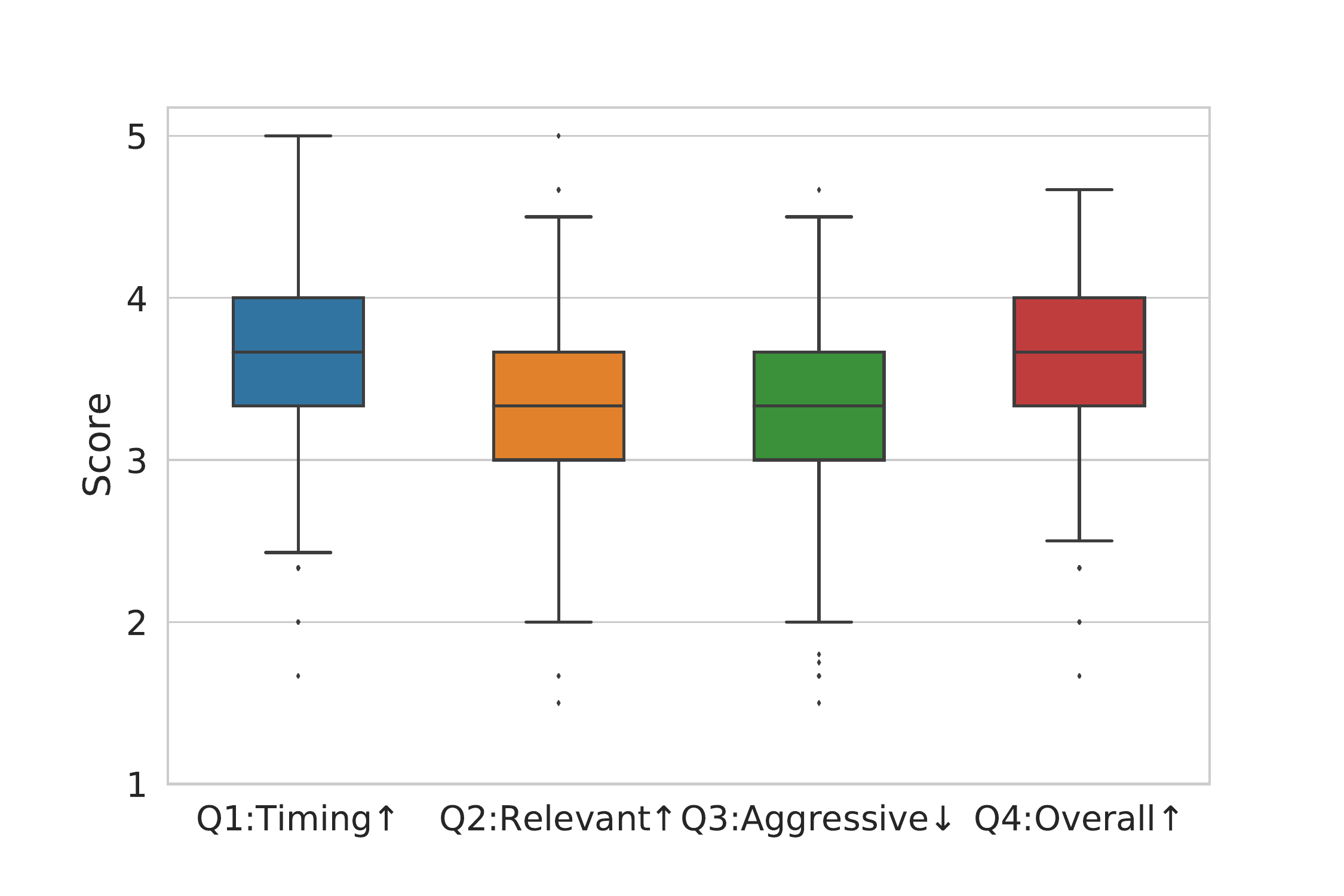}
    \vspace{-2mm}
    \end{minipage}

    \begin{minipage}[b]{.49\linewidth}
    \includegraphics[trim=1cm 1cm 0.5cm 1cm, clip=true,width=1.1\linewidth]{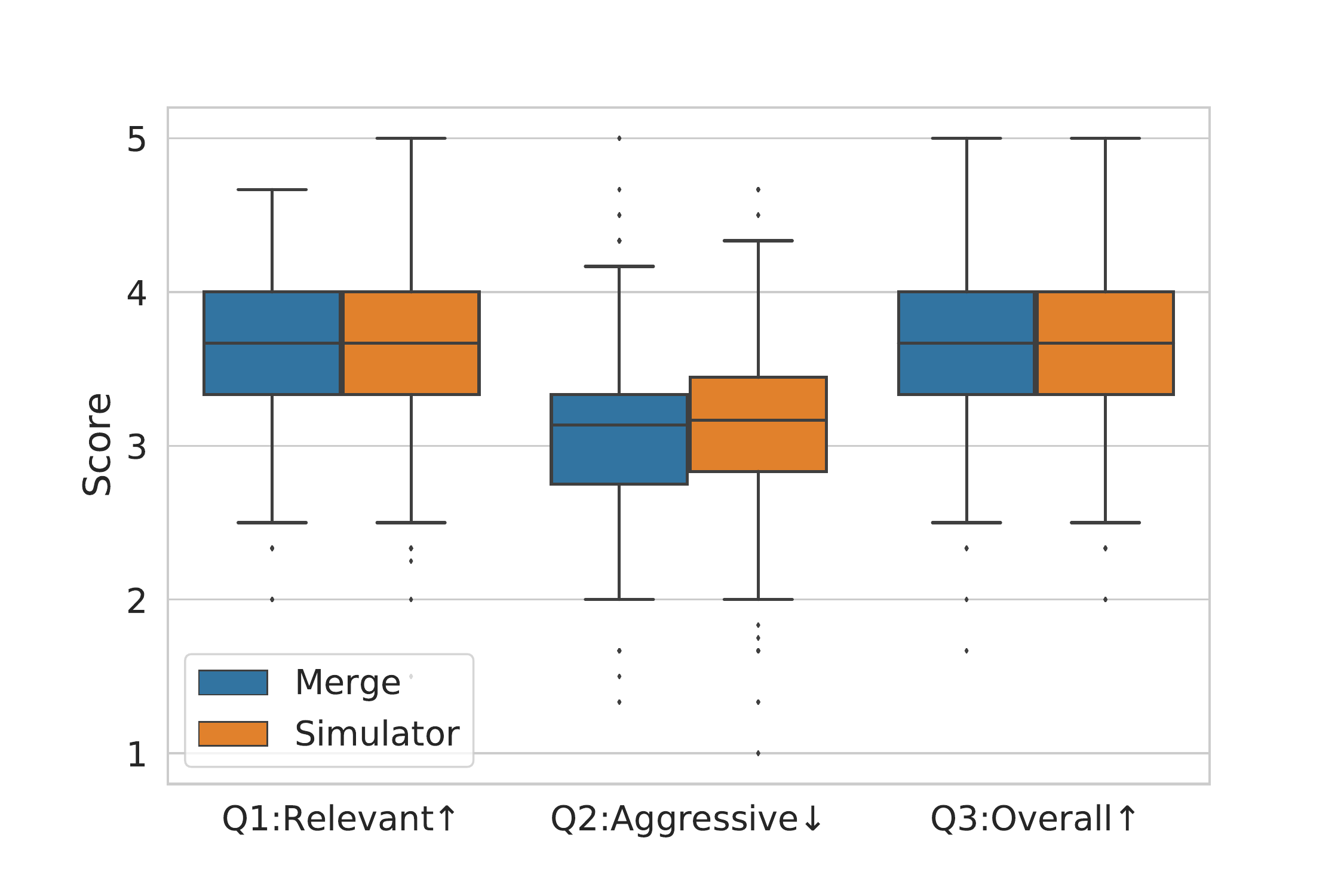}
    \vspace{-2mm}
    \caption*{Task 1: Conversation Evaluation}
    \end{minipage}
    \begin{minipage}[b]{.49\linewidth}
    \includegraphics[trim=1cm 1cm 0.5cm 1.5cm, clip=true,width=1.1\linewidth]{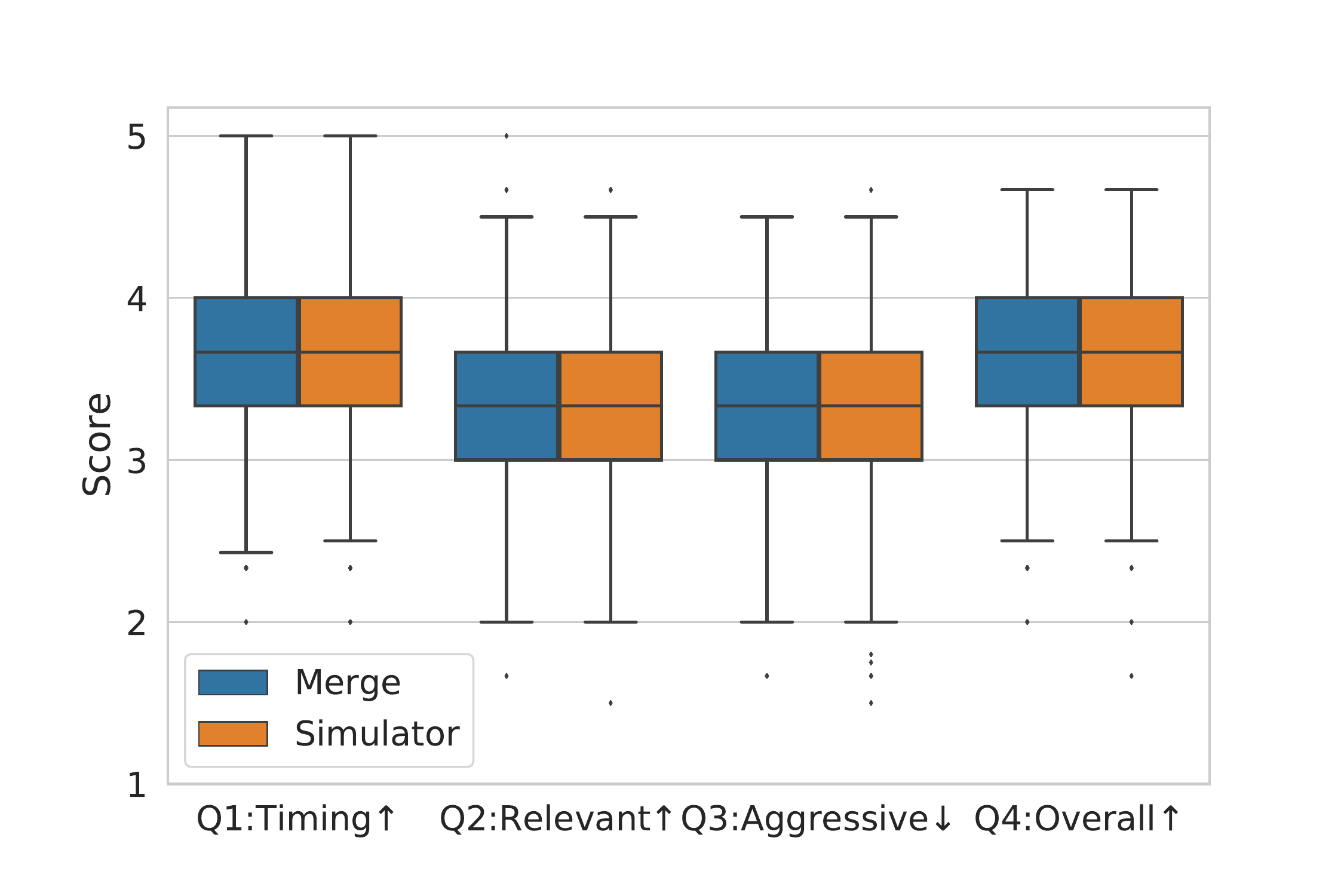}
    \vspace{-2mm}
    \caption*{Task 2: Transition Evaluation}
    \end{minipage}
\vspace{-2mm}
\caption{Score distribution of task 1 (left) and 2 (right). The top charts are  averaged scores over three workers for all dialogues. The bottom charts are the separated averaged scores where TOD is from \textit{Merge SGD} and \textit{Simulators}. $\uparrow$ indicates the higher score the better of this aspect and vice versa for $\downarrow$. }
\label{fig:average_score_box_plot}
\vspace{-3mm}
\end{figure*}

\section{Results and Analysis}
For brevity, 
we use 
\textit{T} to denote \textit{Task} in the following. Each dialogue is evaluated by three crowdworkers so that we can check the annotation variance for reliable results.

\subsection{Generated Dialogue Evaluation}
Table \ref{tab:dataset_statistics} presents the statistics of the randomly sampled dialogues submitted to AMT. The average length of chit-chat turns in \textit{Merge SGD} and \textit{TOD Simulation} are about 4.5.
The evaluation results of all dialogues are visualized in the top charts of Figure~\ref{fig:average_score_box_plot}, and the bottom charts show the results for existing TOD data (Merge) and simulator-generated TOD (Simulator).

\begin{table}[t!]
\small
\centering
\begin{tabular}{lrr}
\hline
\bf Intent           & \bf \#Dialogues & \bf Avg Length \\
\hline\hline
FindMovies       &     1,792     &     18    \\
GetTimesForMovie &     30       &     19    \\
FindAttractions  &     1,296     &     16    \\
LookupMusic      &     490      &     16    \\
PlaySong         &     300      &     15    \\
LookupSong       &     8        &     18    \\
\hline\hline
Merge SGD        &     2,014     &     21    \\
TOD Simulation    &     1,902     &     13    \\
\hline
Total & 3,916 & 17\\
\hline
\end{tabular}
\vspace{-2mm}
\caption{Statistics of the sampled dialogues.}
\label{tab:dataset_statistics}
\vspace{-3mm}
\end{table}

It can be observed that our framework is able to produce \emph{context-relevant} task-oriented conversations to match the topic of open-domain dialogues (Q1 in T1; Q2 in T2). This indicates that we can ensure the dialogue flow from open-domain to task-oriented dialogues is natural. The median relevance scores are slightly higher than the \emph{Neutral} line, suggesting that our sales simulator can perform his sales strategy without annoying customers.
The observation further demonstrates the feasibility and effectiveness of our proposed method. 
In terms of the salesperson's aggressiveness, crowdworkers think that the transition is neutral and somewhat aggressive, showing that smoothly transitioning is still an important research problem to explore.
Furthermore, the transition timing scores (Q1 in T2) also demonstrate that our proposed task-oriented intent detection can capture a suitable moment in a zero-shot setting, so that the sales may not miss any business opportunity of product promotion.

We can observe that most of overall scores (Q3 in T1; Q4 in T2) are above \textit{Neutral} (Score 3)\footnote{The full description of each score is presented in Appendix \ref{sec:guideline}.}, indicating that the generated dialogues and transitions are overall good for a salesperson's business perspective.
The human judgement demonstrates that our proposed approach is capable of simulating a large-scale \emph{reasonable} dialogues aligned with our purpose, implying that both research community and industries can greatly benefit from our released data and the built simulators that can continuously generate more data for training.
Our framework and the constructed dataset reduce the cost for large-scale data requirement for better practice.

\begin{figure}[t]
    \centering
    \includegraphics[trim=0.08cm 0 0 0,clip,width=0.44\textwidth]{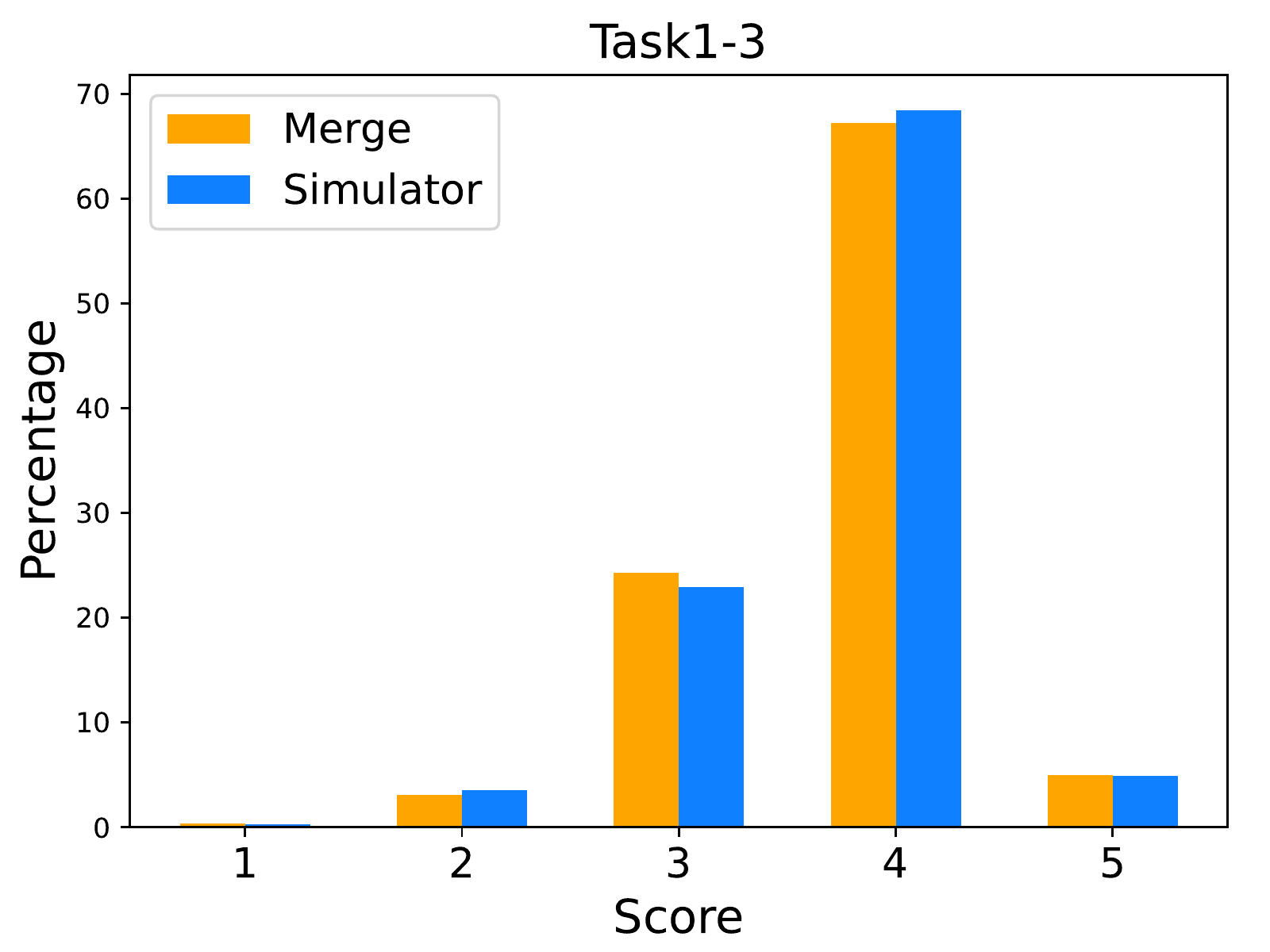}
    \vspace{-2mm}
    \caption{The score distribution between \textit{Merge SGD} and \textit{TOD Simulators} in terms of their overall dialogue quality (Q3 of T1).}
    \label{fig:merge_vs_simulator}
\end{figure}

To further investigate whether the proposed \emph{TOD simulators} described in Section \ref{ssec:dialogue_generation} can generate reasonable dialogues compared to \textit{Merge SGD}, we visualize their individual scores as shown at the bottom of Figure \ref{fig:average_score_box_plot}. 
There is no significant difference between two groups, and we further investigate their score distribution of each question shown in Figure \ref{fig:merge_vs_simulator}. 
Both results tell that given the context of open-domain utterances, our TOD simulators are able to generate the suitable task-oriented dialogues with comparable quality to those from the publicly available benchmark TOD data--SGD. Consequently, our framework can be utilized to generate large-scale data cost-effectively and the generation quality is comparable with the current benchmark dialogue data. 
 
\begin{table}[t!]
\small
\centering
\begin{tabular}{lc}
\hline
\bf Detector & \bf Avg Rank (std.) \\
\hline\hline
\textbf{Detector1}: SQuAD 2.0 & 1.74 $\pm$ 0.48\\
\textbf{Detector2}: + Commonsense data & 1.77 $\pm$ 0.48\\
\textbf{Detector3}: TransferQA & 2.00 $\pm$ 0.52\\
\hline
\end{tabular}
\vspace{-2mm}
\caption{The average ranks of three detectors.}
\label{tab:detector-rank}
\vspace{-3mm}
\end{table}

 
\subsection{Intent Detector Comparison}
Table \ref{tab:detector-rank} shows the average ranks of three detectors described in T3.
We find that Detector1 (pre-trained on SQuAD 2.0) and Detector2 (pre-trained on SQuad 2.0, SWAG, CommonsenseQA) perform almost the same, implying that simply pre-training on extra commonsense-related QA data may not significantly improve the ability of detecting implicit intents.
Possible reasons may be either that these datasets include quite similar knowledge about our target intents, or our zero-shot QA model reaches its capacity bottleneck. 
How to better utilize commonsense knowledge for detecting potential intents can be further investigated in the future.
\citet{lin-etal-2021-zero} has demonstrated Detector3 (trained on several QA datasets) is able to achieve decent dialogue state tracking performance in zero-shot settings. Therefore, we did not fine-tune it on the task-oriented datasets such as SGD Detector1 and Detector2 are fine-tuned on. 
However, according to its average rank, Detector3 is significantly worse than other detectors.
Probably because the intents in chit-chat conversations are more implicit and complex than task-oriented intents, the ability of detecting implicit intents cannot be easily transferred.

\subsection{Potential Research of Released Data}
In addition to the proposed framework and the released dataset, our collected human judgement has the potential of providing valuable contributions to dialogue community and industrial products. Each question along with its corresponding scores can be treated as a interested task, and we briefly describe some (but not limited to) examples of crowdsourced data usage.

The human scores from T1 can be formulated as classification or regression annotations which measure the relevance between a recommended product and a conversation context, whether a salesperson in a dialogue is too aggressive, or the overall quality of a sales dialogue.
Similarly, we can apply these ideas to T2, which focuses on evaluating transitions. Particularly, deciding when is a good to perform a transition can be an interesting topic for future research. This will also benefit industries to develop more intelligent dialogue systems interacting with customers.
Moreover, the rank annotations provided by workers from T3 can be considered as high-quality data for training a ranking model or an intent detector. Apart from this, the data can also be utilized as a gold standard to assess the performance of different algorithms predicting user implicit intents. We expect these examples will inspire the community and industries to discover more interesting research directions and applications. 

\section{Related Work}
Our work is related to dataset construction for building persuasive dialogue systems that try to persuade the participant to take a specific action. \citet{hiraoka-etal-2014-reinforcement} annotated 34 dialogues, in which an experienced salesperson tries to convince a customer to buy a camera. \citet{yoshino-etal-2018-dialogue} requested crowdsourcing workers to generate 200 persuasive dialogues. In each dialogue, one participant persuaded another one to adopt his suggestion such as cleaning a room. \citet{wang-etal-2019-persuasion} collected 1017 dialogues, in which one of the participants was convinced to donate to a specific charity. We can see that the covered conversation scenarios in these datasets were strictly limited to specific tasks, while our scenarios are more general and can be easily extended to different cases.
Also, our constructed dataset is about three times larger than the prior work, indicating the usefulness of the recent pre-trained paradigm.

The topic of conversational recommendation systems is also related to our work.
Similarly, \citet{liu-etal-2020-towards-conversational} also built a dataset by asking human workers to create dialogues based on a topic path.  
It should be noted that, in these datasets, the goal of such systems is to only make \emph{entity} recommendations instead of \emph{tasks}, while our work goes beyond them in naturally transferring from chit-chat to task-oriented dialogues and completing a task the user may want.

Another related work is generating a transition between two given open-domain utterances. \citet{tang-etal-2019-target} proposed to generate the transition  conditional on a specific word, because they want the generated transition can drive the conversation topic to the specified word. \citet{sevegnani-etal-2021-otters} collected a new dataset of human-created one-turn topic transitions. Each dialogue contains 2 utterances with different topics and 1 transition in the middle of them.

There are some recent studies trying to merge chit-chat and task-oriented dialogues, but the purposes of merged dialogues differ from ours. \citet{sun-etal-2021-adding} enhanced the utterances in task-oriented dialogues by appending chit-chat sentences. They hope that the agent gains more social, personalized, and engaging communication skills. \citet{ennen2021universal} proposed a dialogue system that can transfer the style of generated response from chit-chat to task-oriented styles. However, the system is a prototype model, there is still a large gap to properly bridge chitchat and task-oriented dialogues. The motivation of our work is closely similar to the studies by \citet{yu2017learning} and \citet{young2022fusing}. \citet{yu2017learning} manually created several task-oriented response generation strategies specifically designed for the movie promotion scenario. In addition, the expert knowledge was utilized to design reinforcement learning rewards that help their dialogue system to decide which action to take (i.e., continuing chit-chat or selecting a task-oriented strategy to reply). 
In order to fuse open-domain and task-oriented dialogues to a complete and natural conversation, \citet{young2022fusing} manually rewrote existing task-oriented utterances and added new open-domain conversations. The most crucial difference between their work and ours is that, in their dialogues, the user ``explicitly'' expressed his/her intentions indicating clear clues about when and how to naturally transit from chit-chat to task-oriented conversations, while our user intentions are ``implicit'' which makes detection and transition more challenging.

However, we also observe that the prior work in these studies heavily relied on human efforts (data collection, expert-created strategies, etc.).
Therefore, it can be expensive and hard to extend their data or method the practical cases due to the requirement of larger-scale training data.
Our proposed framework benefits from the pre-trained models and shows its outstanding conversational capability.
The flexibility of extending to diverse cases is also validated, considering that all components inside the framework can be easily substituted by the updated models, and the generated data can be used by semi-supervised or unsupervised methods for cold-start scenarios.

\section{Conclusion}
This paper proposes a novel framework to generate dialogues that naturally transition from open-domain to task-oriented scenarios at a large scale without heavy human efforts.
Our proposed chit-chat to task-oriented transition approach can capture the suitable timing when the user shows the implicit intents and generate the diverse and natural transition turn to trigger the task-oriented utterances.
Our human evaluation shows that the automatically generated dialogues have a reasonable quality with natural conversation flows from a business point of view.
The released dataset and framework empowers research community to easily obtain large-scale target dialogues and the human annotated scores can be utilized for related work.
This paper has a great potential of guiding future research directions and benefiting the community of both research and industry.

\section*{Acknowledgements}
We thank reviewers for their insightful comments.
This work was financially supported from MediaTek Research, Amazon AWS Machine Learning Research Awards, and the Young Scholar Fellowship Program by Ministry of Science and Technology (MOST) in Taiwan, under Grant 111-2628-E-002-016.

\bibliography{anthology,custom}
\bibliographystyle{acl_natbib}
\appendix

\section{Questions for Intent Detection}
\label{sec:questions}

\begin{table}[h]
\centering
\small
\begin{tabular}{p{2.35cm} p{4cm}}
\toprule
\textbf{Intent}           & \textbf{Question}                                               \\
\midrule
\textsf{FindMovies}          &   Is the user asking about finding movies? \\
\textsf{GetTimesForMovie}    &   Is the user asking about getting the time for movies?               \\
\textsf{FindAttractions}     &   Is the user asking about finding attractions?   \\
\textsf{LookupMusic}         &   Is the user asking about looking up music?  \\
\textsf{PlaySong}            &   Is the user asking about playing songs? \\
\textsf{LookupSong}          &   Is the user asking about looking up songs?   \\
\bottomrule
\end{tabular}
\caption{Intent-associated questions.}
\end{table}

\section{Crowdsourcing Guideline}
\label{sec:guideline}

\subsection{Task 1:Salesperson-Customer Conversation}
In order to improve the skills to sell more products, a beginner salesperson is learning dialogue strategies by reading prior conversations between customers and other salespeople. This beginner salesperson needs your help to determine if a salesperson used a good dialogue strategy to conduct an effective and strategic sales conversion.

In more detail, you will be presented with one conversation history between a salesperson and a customer. The salesperson may recommend a movie, a song, attractions and so on for the customer. Instead of recommending a product or service to the customer directly, the salesperson wants to make the recommendation more gradually and naturally by starting the conversation with chit-chat.

In this task, you need to rate the conversation from the following 3 aspects:
\begin{compactitem}
    \item How relevant is the recommended product or service to the conversation context?
    \item How aggressive is the salesperson’s communication strategy?
    \item Do you think the sales conversation is overall a good example of making sales recommendations?
\end{compactitem}

\paragraph{Questions}
\begin{itemize}
    \item How relevant is the recommended product or service to the conversation context?
    \begin{compactitem}
        \item 1: Not at all (it is impossible for me to find the relevance between the recommended item and the context)
        \item 2: Less than neutral (it is a bit hard for me to find the relevance between the recommended item and the context)
        \item 3: Neutral (With some effort, I can find a reasonable relevance between the recommended item and the context)
        \item 4: Relevant (I can easily find that the recommend item has obvious relevance with the context, even though the recommended item is not perfectly matching the context)
        \item 5: Very Relevant (the recommended item is perfectly matching the context)
    \end{compactitem}
    \item How aggressive is the salesperson’s communication?
        \begin{compactitem}
             \item 1: Not aggressive at all (the conversation flows very naturally and smoothly from chit-chat to making recommendations; If I was the customer, I feel very comfortable when the salesperson is making recommendations)
            \item 2: Less than neutral (The flow of the conversation is generally natural and smooth, although there are few imperfections)
            \item 3: Neutral (The salesperson starts to recommend an item; It is ok to me)
            \item 4: Aggressive (The salesperson suddenly starts to recommend an item; this makes me a bit uncomfortable)
            \item 5: Very aggressive (The salesperson suddenly starts to recommend an item; this makes me very uncomfortable)
        \end{compactitem}
    \item Is the sales conversation overall a good example to the beginner salesperson?
        \begin{compactitem}
            \item 1: Not at all (This example is really very bad; the beginner salesperson should not spend time on learning this example)
            \item 2: Less than neutral (This example is not good; it would not be a pity if the beginner salesperson skips it)
            \item 3: Neutral (This is not a bad example; the beginner salesperson may learn some useful dialogue skills from it, but not very much)
            \item 4: Good (This is a good example of making recommendations; the imperfections can be ignored; the beginner salesperson should keep this example in his mind)
            \item 5: Very good (This is a perfect example of making recommendations; the beginner salesperson should keep it deeply in his mind)
        \end{compactitem}
\end{itemize}

\subsection{Task 2: Chit-Chat to Task-Oriented Transition}
In order to improve the skills to sell more products, a beginner salesperson is learning dialogue strategies by reading prior conversations between customers and other salespeople. This beginner salesperson needs your help to determine if a salesperson used a good dialogue strategy to conduct an effective and strategic sales conversion.

You will be presented with a conversation between a salesperson and a customer. The salesperson may recommend a movie, a song, attractions and so on for the customer. Instead of recommending a product or service to the customer directly, the salesperson wants to make the recommendation more gradually and naturally by starting the conversation with chit-chat. Once the salesperson thinks it is the right time, he will say something (named \textbf{transition} in this task) to change the conversation from chit-chat to recommendation-making. 

In this task, you will need to rate the transition from the following 4 aspects:
\begin{itemize}
    \item Is it the right time to make the transition?
    \item Is the transition relevant to the conversation context?
    \item Is the transition aggressive?
    \item Is the transition overall good?
\end{itemize}

\paragraph{Questions}
\begin{itemize}
    \item Is it the right time to make the transition?
    \begin{itemize}
        \item 1: Very bad time (This is definitely not the right time to do it. It is highly likely that the customer will find you very annoying)
        \item 2: Bad time (This is not a good time to make the transition. It may cause negative customer feelings)
        \item 3: Neutral (I don’t think making the transition at the time is good, but it is ok to me to continue the conversation if I was the customer)
        \item 4: Good time (it is a good time to make the transition, but maybe it will be perfect if the transition is made earlier or later)
        \item 5: Very good time (it is a perfect time to make the transition)
    \end{itemize}
    \item Is the transition relevant to the conversation context?
    \begin{itemize}
        \item 1: Not at all (it is impossible for me to find the relevance between the transition and the context)
        \item 2: Less than neutral (it is a bit hard for me to find the relevance between the transition and the context)
        \item 3: Neutral (With some effort, I can find a reasonable relevance between the transition and the context)
        \item 4: Relevant (I can easily find that the transition has obvious relevance with the context, even though the transition is not perfectly matching the context)
        \item 5: Very Relevant (the transition is perfectly matching the context; it is hard for me to find a better transition)
    \end{itemize}
    \item Is the transition aggressive?
    \begin{itemize}
        \item 1: Not aggressive at all (the conversation flows very naturally and smoothly from chit-chat to making the transition; If I was the customer, I feel very comfortable when the salesperson is doing it)
        \item 2: Less than neutral (The flow of the conversation is generally natural and smooth, although there are few imperfections)
        \item 3: Neutral (The salesperson starts to make the transition; It is ok to me)
        \item 4: Aggressive (The salesperson suddenly starts to make the transition; this makes me a bit uncomfortable)
        \item 5: Very aggressive (The salesperson suddenly starts to make the transition; this makes me very uncomfortable)
    \end{itemize}
    \item Is the transition overall good?
    \begin{itemize}
        \item 1: Not at all (This transition is really very bad; the beginner should not spend time on leaning this transition)
        \item 2: Less than neutral (This transition is not good; It would not be a pity if the beginner salesperson skips this example)
        \item 3: Neutral (This is not a bad transition; the beginner salesperson may learn some useful dialogue skills from it, but not very much)
        \item 4: Good (This is a good example of making a transition; the imperfections can be ignored; the beginner salesperson should keep this example in his mind)
        \item 5: Very good (This is a perfect example of making a transition; the beginner salesperson should keep it deeply in his mind)
    \end{itemize}
    \item Which transition of the following do you think is the best?
    \begin{itemize}
        \item \textit{transition 1}
        \item \textit{transition 2}
        \item \textit{transition 3}
        \item \textit{transition 4}
    \end{itemize}
\end{itemize}

\subsection{Task 3: Customer's Implicit Intent}
In order to improve skills to sell more products, some beginner salespersons are practicing dialogue strategies by reading prior conversations between customers and other salespeople. When reading a conversation, they will try to guess what the customer is thinking or what the customer might be most likely interested in. These beginner salespersons need your opinions about the reasonability of their answers.

In this task, you will be presented with a conversation snippet between a salesperson and a customer. These beginners provided their guesses right after a customer's utterance. There are three sets of intent detected by different salespersons. You will need to rank them in terms of the intent relevance (implicit intent) with the conversation. If they have the exactly same intent, you can give them the same rank. Otherwise, please decide which is the better one. 1 for the best intents. 3 for the worst intents. In addition, "None" means there isn't any intent detected by the salespersons.

\noindent \textbf{Example}

\begin{table}[h]
\small
    \begin{tabular} { | p{0.7cm} p{6.0cm} | }
     \hline
     \textbf{Sales:} & Hello, what is your hobby?\\ 
     \textbf{User:} & I like to read a lot. I also like to go to the movies. What about yourself? \\ 
     &\\
     & $\cdot$ [FindMovies, LookupMusic] \\
     & $\cdot$ [PlaySong] \\
     & $\cdot$ [LookupMusic] \\
     \hline
    \end{tabular}
    \vspace{-1mm}
\vspace{-2mm}
\end{table}

All possible intents might exist in the conversation.
\begin{itemize}
\small
    \item LookupSong: find songs to listen to
    \item PlaySong: play songs
    \item LookupMusic: find music to listen to
    \item FindMovies: find movies to watch
    \item GetTimesForMovie: obtain the available time for watching a movie
    \item FindAttractions: find attractions to visit
\end{itemize}

\paragraph{Questions}
\begin{itemize}
    \item Please select the rank for the [FindMovies, LookupMusic].
    \begin{itemize}
        \item 1
        \item 2
        \item 3
    \end{itemize}
    \item Please select the rank for the [PlaySong].
    \begin{itemize}
        \item 1
        \item 2
        \item 3
    \end{itemize}
    \item  Please select the rank for the [LookupMusic].
     \begin{itemize}
        \item 1
        \item 2
        \item 3
    \end{itemize}
    \item  Please choose your own answers given this conversation.
    \begin{itemize}
        \item LookupSong
        \item PlaySong
        \item LookupMusic
        \item FindMovies
        \item GetTimesForMovie
        \item FindAttractions
        \item None
    \end{itemize}
\end{itemize}

\clearpage

\section{Generated Dialogue Samples}
\newblock

\begin{tabular}{lp{0.8\textwidth}}
\hline
\hline
\multicolumn{2}{c}{\textbf{Task-oriented Simulators}} \\ [1ex]
\hline
\textbf{Sales: } & Oh, I'm sorry to hear that. Do you have any other family members that you are close to? \\
\textbf{User: } & Yes, I have two brothers and two sisters. We are all very close. How about you? \\
\textbf{Sales: } & I don't have any siblings, but I have a lot of cousins that I'm very close to. \\
\textbf{User: } & That's great. I'm glad you have a good relationship with your cousins. What do you like to do for fun? \\
\textbf{Sales: } & I like to play video games and watch movies. What about you, what do you enjoy doing? \\
\textbf{User: } & I like playing video games as well. I also like to go to the movies. I love going to the theater. - \underline{\textbf{FindMovies}} \\
\textbf{Sales: } & \textit{Sure, what kind of movies do you like to watch? Can you see [movie\_name]? - [Transition]} \\
\textbf{User: } & I would love to see Captain Marvel. I'd like to watch it at Century at Tanforan. \\
\textbf{Sales: } & I found [count] movie you might like. How about [movie\_name]? \\
\textbf{User: } & Yeah, Hotel Mumbai sounds good. \\
\textbf{Sales: } & You want to play [title] without subtitles, is that right? \\
\textbf{User: } & Yes, that's right. \\
\textbf{Sales: } & Okay, so you want me to play the movie you requested. I'll start the movie now. \\
\textbf{User: } & Yes, please do. \\
\textbf{Sales: } & Can you please confirm that you want to play [title] without subtitles? \\
\hline
\hline
\end{tabular}

\vspace*{0.6 cm}

\begin{tabular}{lp{0.8\textwidth}}
\hline
\hline
\multicolumn{2}{c}{\textbf{Merge SGD}} \\ [1ex]
\hline
\textbf{User: } & I like to drive around in my black car and listen to music. What do you like to do? \\
\textbf{Sales: } & I love going to the beach and swimming. What kind of music do you listen to? \\
\textbf{User: } & I listen to a lot of rap and hip hop. Do you have a favorite rapper? - \underline{\textbf{LookupMusic}} \\
\textbf{Sales: } & \textit{I’m a top notch rapper and I’m a rock musician. Are you in the mood for the music. - [Transition]} \\
\textbf{User: } & I'm in the mood for some music. Can you find songs from the album Camila. \\
\textbf{Sales: } & What about the song [song\_name] from the album [album] by [artist]? \\
\textbf{User: } & I'm not in the mood for that one, do you have a different song? \\
\textbf{Sales: } & What about the song [song\_name] from the album [album] by [artist]? \\
\textbf{User: } & What is the genre of the song? \\
\textbf{Sales: } & Real Friends is a [genre] song. \\
\textbf{User: } & Sounds good. \\
\textbf{Sales: } & Do you want to hear the song now? \\
\textbf{User: } & Yes, I would love to. \\
\textbf{Sales: } & You want to play [song\_name] on your [playback\_device]? \\
\textbf{User: } & No, I would prefer to play it on my kitchen speaker. \\
\textbf{Sales: } & You want to play [song\_name] on your [playback\_device]? \\
\textbf{User: } & Yes, that's right. \\
\textbf{Sales: } & The song is now playing. \\
\textbf{User: } & Thank you very much. \\
\textbf{Sales: } & Is there anything else I can help you with? \\
\textbf{User: } & No, that is all. Thank you for your time. \\
\textbf{Sales: } & Enjoy your music. Have a wonderful day. \\
\hline
\hline
\end{tabular}

\begin{table*}[t]
\begin{tabular}{lp{0.8\textwidth}}
\hline
\hline
\multicolumn{2}{c}{\textbf{Merge SGD}} \\ [1ex]
\hline
\textbf{User: } & London is a great place to visit, but I wouldn't want to live there. Too many people.  \\
\textbf{Sales: } & I know what you mean. London is the most populous city in the United Kingdom.  \\
\textbf{User: } & It's also the most densely populated metropolitan area in the world.  \\
\textbf{Sales: } & Yes, it is. It has a population of 8,537,673. \\
\textbf{User: } & Wow, that's a lot of people. Do you live in the city or the suburbs?  \\
\textbf{Sales: } & I live in London, the capital of the UK. It's a great city to live in.  \\
\textbf{User: } & I've never been there but I've always wanted to go. I've heard it's the cultural, financial, and media capital of Europe.  \\
\textbf{Sales: } & Yes, that is true. It is also one of the world's most visited cities as well.  \\
\textbf{User: } & I would love to go there and see all of the museums and historical sights. -\underline{\textbf{FindAttractions}} \\
\textbf{Sales: } & \textit{Sure. What city? What kind of attraction are you looking for? - [Transition]} \\
\textbf{User: } & I'm looking for attractions to visit  \\
\textbf{Sales: } & In which city do you want to search for?  \\
\textbf{User: } & I would like attractions in San Francisco  \\
\textbf{Sales: } & I found [count] attractions, for example [attraction\_name], a [category] \\
\textbf{User: } & Is it a child-friendly place? Is entry free? \\
\textbf{Sales: } & The entry is free and it's a good place for kids \\
\textbf{User: } & Very good! Can you give me any other suggestions? Any attraction is fine \\
\textbf{Sales: } & I found [count] attractions. I suggest [attraction\_name], a [category] \\
\textbf{User: } & This is fine too \\
\textbf{Sales: } & You need anything else? \\
\textbf{User: } & Yes, I'd like to watch a movie there, preferably a 3d show \\
\textbf{Sales: } & There are [movie\_name], [movie\_name] and [movie\_name]. \\
\textbf{User: } & I'd like other proposals \\
\textbf{Sales: } & There are [movie\_name], [movie\_name] and [movie\_name] \\
\textbf{User: } & I like IT two! What online movies do you propose? \\
\textbf{Sales: } & What kind of movie do you prefer? \\
\textbf{User: } & I love Suspense movies \\
\textbf{Sales: } & I found [count] movies: do you prefer [title], [title] or [title]? \\
\textbf{User: } & I prefer The Art of Self-Defense! I want to watch it now with English subtitles \\
\textbf{Sales: } & Playing [title] with [subtitle\_language] subtitles: do you confirm? \\
\textbf{User: } & No, I prefer with French subtitles, thanks! \\
\textbf{Sales: } & Playing [title] with [subtitle\_language] subtitles: do you confirm? \\
\textbf{User: } & Now it's correct, I confirm. \\
\textbf{Sales: } & The movie is starting! \\
\textbf{User: } & Who is the main actor? \\
\textbf{Sales: } & The mai actor is [starring]. \\
\textbf{User: } & Thank you! It's all for now! \\
\hline
\hline
\end{tabular}

\caption{These are more examples generated by different strategies in our framework. The text has the \underline{underline} is the detected intent. The sentence in \textit{italics} is the transition.}
\end{table*}



\end{document}